\documentclass[10pt,twocolumn,letterpaper]{article}

\usepackage[pagenumbers]{iccv} 

\usepackage{adjustbox}
\usepackage{booktabs}
\usepackage{graphicx}
\usepackage{amssymb}
\usepackage{amsmath}
\usepackage{multirow}
\usepackage{algorithm}
\usepackage{algpseudocode}
\usepackage{algorithmicx}
\usepackage{float}
\usepackage{xcolor}
\usepackage[accsupp]{axessibility}
\usepackage{nicefrac}

\definecolor{iccvblue}{rgb}{0.21,0.49,0.74}
\usepackage[pagebackref,breaklinks,colorlinks,allcolors=iccvblue]{hyperref}

\title{Gradient Extrapolation for Debiased Representation Learning}

\author{Ihab Asaad$^1$ \quad Maha Shadaydeh$^1$\quad 
Joachim Denzler$^1$ \\
$^1$Computer Vision Group, Friedrich Schiller University Jena, Germany\\
{\tt\small \{ihab.asaad, maha.shadaydeh, joachim.denzler\}@uni-jena.de}}

\begin{document}
\maketitle
\begin{abstract}
Machine learning classification models trained with empirical risk minimization (ERM) often inadvertently rely on spurious correlations. When absent in the test data, these unintended associations between non-target attributes and target labels lead to poor generalization.
This paper addresses this problem from a model optimization perspective and proposes a novel method, \underline{G}radient \underline{E}xtrapolation for Debiased \underline{R}epresentatio\underline{n} L\underline{e}arning (GERNE), designed to learn debiased representations in both known and unknown attribute training cases. 
GERNE uses two distinct batches with different amounts of spurious correlations and defines the target gradient as a linear extrapolation of the gradients computed from each batch's loss. Our analysis shows that when the extrapolated gradient points toward the batch gradient with fewer spurious correlations, it effectively guides training toward learning a debiased model.
GERNE serves as a general framework for debiasing, encompassing ERM and Resampling methods as special cases. 
We derive the theoretical upper and lower bounds of the extrapolation factor employed by GERNE. By tuning this factor, GERNE can adapt to maximize either Group-Balanced Accuracy (GBA) or Worst-Group Accuracy (WGA). 
We validate GERNE on five vision and one NLP benchmarks, demonstrating competitive and often superior performance compared to state-of-the-art baselines. The project page is available at: \url{https://gerne-debias.github.io/}.
\end{abstract}
    
\section{Introduction}
\label{sec:intro}
\begin{figure*}[ht]
    \centering
    \begin{subfigure}[b]{0.34\textwidth}
        \centering
        \includegraphics[width=\textwidth,trim={0cm 0.15cm 0 0.1cm},clip]{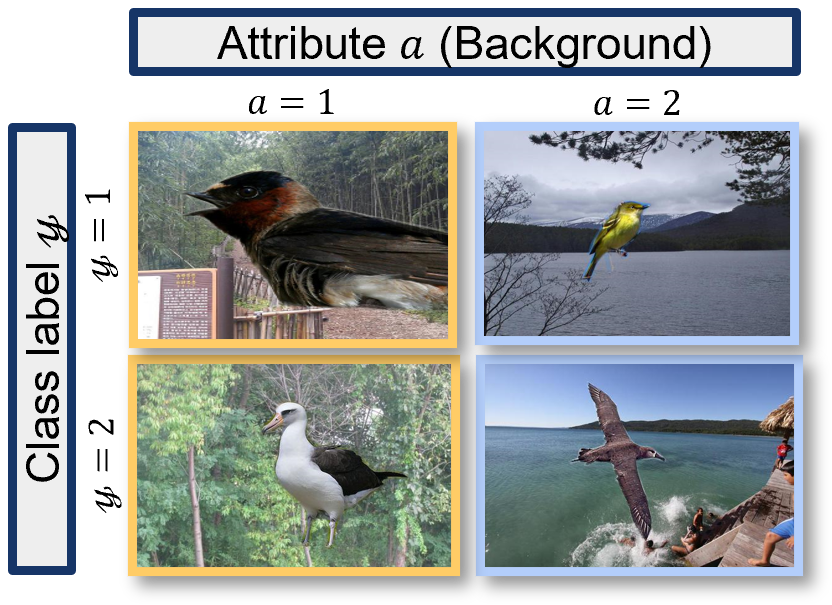}
        \caption{}
        \label{fig:subfig1}
    \end{subfigure}
    \hfill
    \begin{subfigure}[b]{0.34\textwidth}
        \centering
        \includegraphics[width=\textwidth,trim={0.cm 0 0 0.05cm},clip]{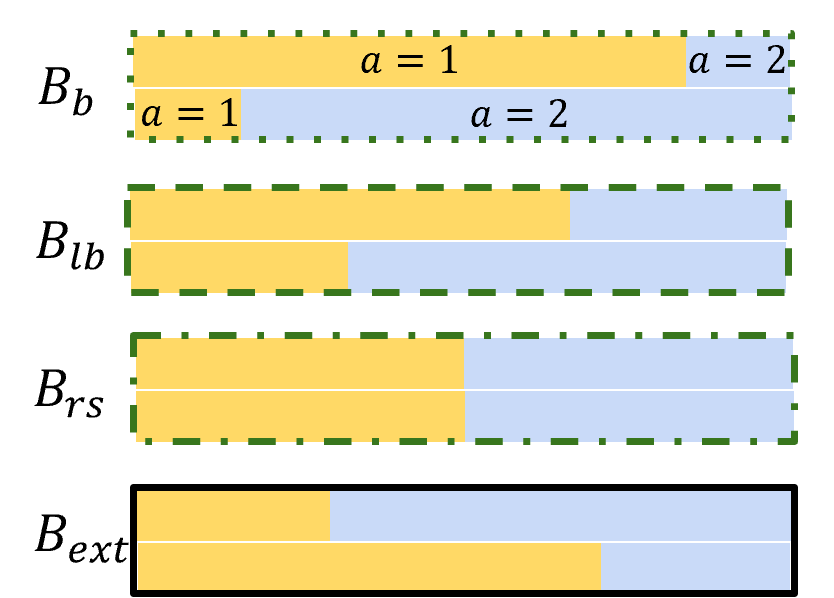}
        \caption{}
        \label{fig:subfig2}
    \end{subfigure}
    \hfill
    \begin{subfigure}[b]{0.23\textwidth}
        \centering
        \includegraphics[width=\textwidth,trim={0cm 0 0.65cm 0.cm},clip]{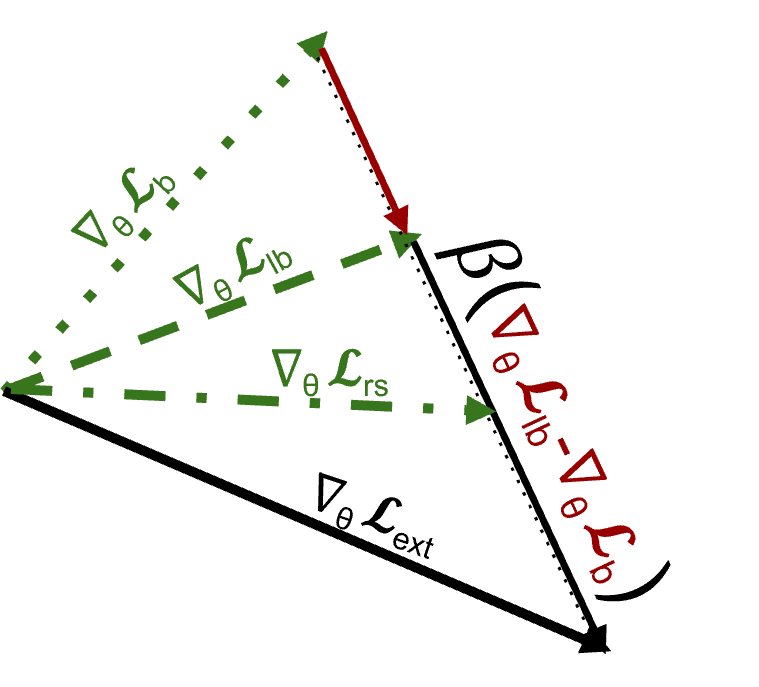}
        \caption{}
        \label{fig:subfig3}
    \end{subfigure}
    \caption{
    \subref{fig:subfig1}: Sample images from the waterbirds classification task. Most landbird images appear with land backgrounds (i.e., $y=1, a=1$), while most waterbird images appear with water backgrounds (i.e., $y=2, a =2$). This correlation between bird class and background introduces spurious correlations in the dataset.
    \subref{fig:subfig2}: Visualization of batch construction. $B_b$ refers to a biased batch where the majority of images from class $y=1$ (top row) have attribute $a=1$ (yellow), and the majority of images from class $y=2$ (bottom row) have attribute $a=2$ (light-blue). $B_{lb}$ represents a less biased batch, with a more balanced attribute distribution within each class, controlled by $c$ (here $c=\frac{1}{2}$). $B_{rs}$ depicts a batch with group-balanced distribution and refers to the batch used in the Resampling method~\cite{idrissi2022simple}. $B_{ext}$ simulates GERNE's batch with $c \cdot (\beta+1) > 1$, where the dataset’s minority groups appear as majorities in the batch.
    \subref{fig:subfig3}: A simplified 2D representation of gradient extrapolation where $\theta \in \mathbb{R}^2$.  $\nabla_{\theta} \mathcal{L}_{b}$ is the gradient computed on $B_b$; training with this gradient is equivalent to training with the ERM objective. $\nabla_{\theta} \mathcal{L}_{lb}$ represents the gradient computed on $B_{lb}$. $\nabla_{\theta} \mathcal{L}_{rs}$ is the gradient computed on $B_{rs}$, which is equivalent, in expectation, to an extrapolated gradient with $c\cdot (\beta+1) = 1$. Finally, $\nabla_{\theta} \mathcal{L}_{ext}$ is our extrapolated gradient, with the extrapolation factor $\beta$ modulating the degree of debiasing based on the strength of spurious correlations present in the dataset.}
    \label{fig:GERNE}
\end{figure*}

Deep learning models have demonstrated significant success in various classification tasks, but their performance is often compromised by datasets containing prevalent spurious correlations in the majority of samples ~\cite{ye2024spurious, liu_avoiding_2023, DFR, hashimoto2018fairness}. 
Spurious correlations refer to unintended associations between easy-to-learn, non-target attributes and target labels. These correlations often cause models trained with Empirical Risk Minimization (ERM)~\cite{ERM} to rely on these correlations instead of the true, intrinsic features of the classes ~\cite{geirhos2020shortcut, du2023shortcut, sagawa2020investigation}. 
This occurs because the ERM objective optimizes for the average performance ~\cite{ERM}, thereby biasing the model toward the easy-to-learn features that are predictive for the majority of training samples. As a result, ERM-trained models often exhibit poor generalization when these spurious features are absent in the test data.
For instance, in the Waterbirds classification task ~\cite{waterbirds}, where the goal is to classify a bird as either a waterbird or a landbird, the majority of waterbirds are associated with water backgrounds. In contrast, the majority of landbirds are associated with land backgrounds. A model trained with ERM might learn to classify the birds based on the background---water for waterbirds and land for landbirds---rather than focusing on the birds' intrinsic characteristics.
This reliance on the spurious feature allows the model to perform well on the majority of training samples, where these correlations hold, but fails to generalize to test samples where these correlations are absent (e.g., waterbirds on land). Examples of Waterbirds images shown in \cref{fig:subfig1}. Avoiding spurious correlations is crucial across various applications, including medical imaging ~\cite{bias_medical, fairness_medical}, 
finance ~\cite{finance}, and climate modeling ~\cite{climate}.

This pervasive challenge has spurred extensive research into strategies for mitigating the negative effects of spurious correlations, particularly under varying levels of attribute information availability.
The authors of ~\cite{ChangeIsHard} provide a comprehensive review of the methods and research directions aimed at addressing this issue. 
In an ideal scenario, where attribute information is available in both the training and validation sets, methods can leverage this information to counteract spurious correlations ~\cite{idrissi2022simple, DRO, wu2023discover}. 
When attribute information is available only in the validation set, methods either incorporate this set into the training process ~\cite{DFR, nam2022spread, sohoni_barack_2022} or restrict its use to model selection and hyperparameter tuning ~\cite{LfF, JTT, liu_avoiding_2023, lee_learning_2021, qiu_simple_2023}. 
Despite these efforts, existing methods still struggle to fully avoid learning spurious correlations, especially when the number of samples without spurious correlations is very limited in the training dataset, leading to poor generalization on the test data where these correlations are absent.

In this paper, we adopt a different research approach, seeking to address the issue of spurious correlations from a model optimization perspective. We propose a novel method, Gradient Extrapolation for Debiased Representation Learning (GERNE), to avoid reliance on spurious features and learn debiased representations. 
The core idea is to sample two types of batches with varying amounts of spurious correlations (\cref{fig:subfig2}) and compute the two losses on these two batches.
We assume that the difference between the gradients of these losses captures a debiasing direction. Therefore, we define our target gradient as the linear extrapolation of these two gradients toward the gradient of the batch with fewer amount of spurious correlations (\cref{fig:subfig3}). 
The contributions of this paper can be summarized as follows:
\begin{itemize}
\item We propose GERNE as a general framework for debiasing, with methods such as ERM and Resampling being shown as special cases.
\item We derive the theoretical upper and lower bounds of the extrapolation factor and establish a direct connection between the extrapolation factor and the risk for the worst-case group. We show that tuning this factor within these bounds enables GERNE to adaptively optimize for either Group-Balanced Accuracy or Worst-Group Accuracy.
\item We validate our approach on six benchmarks spanning both vision and NLP tasks, under both known and unknown attribute cases, demonstrating competitive and often superior performance compared to state-of-the-art methods---particularly in scenarios where samples without spurious correlations are scarce.

\end{itemize}

\section{Related Work}
\label{sec:related_work}
\paragraph{Debiasing according to attribute annotations availability.}
Numerous studies have leveraged attribute annotations to mitigate spurious correlations and learning debiased representation \cite{DRO, zhou2021examining, arjovsky2020invariantriskminimization, LISA, Mixup}. 
For instance, Group DRO \cite{DRO} optimizes model performance on the worst-case group by directly minimizing worst-group loss during training. 
While effective, such methods rely on complete attribute annotations, which are often costly and labor-intensive to obtain.
Consequently, recent works have explored approaches that reduce reliance on full annotations by using limited attribute information \cite{DFR, sohoni_barack_2022, nam2022spread}. 
For example, DFR \cite{DFR} enhances robustness by using a small, group-balanced validation set with attribute information to retrain the final layer of a pre-trained model. 
In cases where attribute information is only available for model selection and hyperparameter tuning \cite{creager2021environment, idrissi2022simple, JTT, LfF, zhang_correct-n-contrast_2022}, an initial or auxiliary ERM-trained model is often used to infer the attributes by partitioning the training data into majority and minority groups. Samples on which the model incurs relatively low loss (i.e., high-confidence predictions) are treated as ``easy'' examples---where spurious correlations are likely to hold--- and these examples form the majority group. Conversely, high-loss samples are considered ``hard'' examples, and typically form the minority group where such correlations may not apply \cite{yang2023mitigating}.
This process effectively creates ``easy'' and ``hard'' pseudo-attributes within each class, allowing debiasing methods that traditionally rely on attribute information to be applied. For example, JTT \cite{JTT} first trains a standard ERM model and then trains a second model by upweighting the misclassified training examples detected by the first model. 
Finally, a more realistic and challenging scenario arises when attribute information is entirely unavailable \cite{masktune, ula}---not accessible for training, model selection, or hyperparameter tuning---requiring models to generalize without explicit guidance on non-causal features \cite{ChangeIsHard}.
\paragraph{Debiasing via balancing techniques.} A prominent family of solutions to mitigate spurious correlations across the aforementioned scenarios of annotation availability involves data balancing techniques \cite{ReWeightResample, idrissi2022simple, wang2022causal, sagawa2020investigation, BSoftmax, CBLoss, ReWeightCRT}. These methods are valued for their simplicity and adaptability, as they are typically faster to train and do not require additional hyperparameters. Resampling underrepresented groups to ensure a more balanced distribution of samples \cite{idrissi2022simple, ReWeightResample} or modifying the loss function to adjust for imbalances \cite{ross2017focal} are common examples of these techniques. We demonstrate in \cref{sec:balance} that although the balancing techniques are effective, their performance is constrained in the presence of spurious correlations. In contrast, our proposed debiasing approach mitigates the negative effects of spurious correlations by guiding the learning process in a debiasing direction, proving to be more effective.

\section{Problem Setup}
We consider a standard multi-class classification problem with \( K \) classes and \( A \) attributes. Each input sample \( x_i \in \mathcal{X} = \{x_j \mid j = 1, \dots, N\} \) is associated with a class label $ y_i \in \mathcal{Y} = \{1, \dots, K\} $ and an attribute $a_i \in \mathcal{A} = \{1, \dots, A\}$, where \( N \) is the total number of samples in the dataset.
We define a group $\mathcal{X}_{y, a}$ for $(y, a) \in \mathcal{G} = \mathcal{Y} \times \mathcal{A} $ as the set of input samples $x_i$  with class label $ y$ and attribute $ a$, resulting in $|\mathcal{G}| = K\cdot A$ groups. 
For each class \(y\), we denote by \(\mathcal{X}_y = \bigcup_{a \in \mathcal{A}} \mathcal{X}_{y,a}\) the set of all samples with label \(y\). We assume all groups are non-empty, i.e., \(\forall (y,a) \in \mathcal{G}, \mathcal{X}_{y,a} \neq \emptyset\), and denote the cardinality of any group \(\mathcal{X}_m\) by \(|\mathcal{X}_m|\).

Our goal is to learn the intrinsic features that define the labels, rather than spurious features present in a biased dataset, where spurious correlations are prevalent. 
This would ensure robust generalization when spurious correlations are absent in the test distribution.
Following ~\cite{DRO}, we aim  to learn a function parameterized by a neural network $f^*: \mathcal{X} \rightarrow \mathbb{R}^K$ to minimize the risk for the worst-case group:
\begin{equation}
\label{eq:worst_loss}
    f^* = \arg \min_f \max_{g \in \mathcal{G}} \mathbb{E}_{x \sim p(x|(y,a) = g)} \left[ \ell(y, f(x))\right],
\end{equation}
where $ \ell(y, f(x)) \rightarrow \mathbb{R}$ is the loss function.
\section{The Proposed Method: GERNE}
We build GERNE with the goal of mitigating the impact of spurious correlations.
The core idea of GERNE is to sample two batches with different amounts of spurious correlations, hereafter named the biased batch \( \boldsymbol{B_{b}} \) and the less biased batch \( \boldsymbol{B_{lb}} \) (\cref{fig:subfig2}). Let $\mathcal{L}_{b}, \mathcal{L}_{lb}$ be the losses calculated on \( \boldsymbol{B_{b}} \) and \( \boldsymbol{B_{lb}} \), respectively. We assume that extrapolating the gradients of these two losses towards the gradient of $\mathcal{L}_{lb}$ guides the model toward debiasing as illustrated in \cref{fig:subfig3}.
We first present GERNE for training with known attributes and then generalize GERNE to the unknown attribute case. 
\subsection{GERNE for the  Known Attributes Case}
In the following, we denote by $p(y,a)$ the joint distribution of class label $y$ and attribute $a$ in a sampled batch. During training, we construct two types of batches with different conditional attribute distributions $p(a|y)$: the \textit{biased} and the \textit{less biased} batches. 
Our method defines the target loss as a linear extrapolation between the losses computed on these two batches. A simplified illustration is shown in \cref{fig:GERNE}. Finally, we derive the link between the extrapolation factor and the risk for the worst-case group in \cref{eq:worst_loss}, and theoretically define the upper and lower bounds of this factor.
\subsubsection{Sampling the biased and the less biased batches}
\label{sec:batches_def}
The biased batch and the less biased batches are sampled to satisfy the following two conditions: 
\begin{enumerate}
     \item  Uniform sampling from classes, i.e., $\forall y \in \mathcal{Y}, p(y) = \frac{1}{K}.$
     \item   Uniform sampling from groups, i.e.,   $\forall (y,a) \in \mathcal{G}, p(x|y, a) = \frac{1}{|\mathcal{X}_{y, a}|} \text{ for } x \in \mathcal{X}_{y, a}$.
\end{enumerate}

The \textbf{biased batch (\( \boldsymbol{B_{b}} \))} is sampled with a conditional attribute distribution  $ p_{b}(a|y)$ within each class $y $ to reflect the inherent bias present in the dataset. Specifically, $ p_{b}(a|y)  = \alpha_{ya}$, where:
\begin{equation}
\label{eq:alphayj}
    \alpha_{ya} = \frac{|\mathcal{X}_{y, a}|}{|\mathcal{X}_{y}|}.
\end{equation}
Note that to sample a biased batch, no access to the attributes is required, and uniformly sampling from $\mathcal{X}_y$ for each label $y$ satisfies \cref{eq:alphayj}. The \textbf{less biased batch (\( \boldsymbol{B_{lb}} \))} is sampled with a conditional attribute distribution, denoted as $p_{lb}(a|y)$, which satisfies the following:
$\forall (y,a) \in \mathcal{G}$:
\begin{align}
\label{eq:inequality_p_a_y}
  \ \min(\frac{1}{A}, p_b(a|y))  \leq p_{lb}(a|y) \leq \max(\frac{1}{A}, p_{b}(a|y)).
\end{align}
That is, $B_{lb}$ exhibits a more balanced group distribution than $B_b$, and $\mathcal{L}_{lb}$ quantifies the loss when spurious correlations are reduced in the sampled batch. 
Choosing
\begin{equation}
\label{eq:rel_ps}
p_{lb}(a|y) = (1 - c)\cdot p_b(a|y) + c\cdot \frac{1}{A} = \alpha_{ya} + c \cdot (\frac{1}{A} - \alpha_{ya})
\end{equation}
satisfies the inequality in \cref{eq:inequality_p_a_y}, where \( c \in (0,1] \) is a hyperparameter that controls the degree of bias reduction. An example of the two types of batches is presented in \cref{fig:subfig2}.
\subsubsection{Gradient extrapolation}
\label{sec:loss_fn}
We define our target loss $\mathcal{L}_{ext}$ as follows:
\begin{equation}
\label{eq:relation_losses}
   \mathcal{L}_{ext} = \mathcal{L}_{lb} + \beta\cdot(\mathcal{L}_{lb}-\mathcal{L}_b),
\end{equation} 
where $\beta$ is a hyperparameter, and the loss form given the joint distribution $p(x, y, a) $ is defined as: 
\begin{equation}
\label{eq:general_loss}
    \mathcal{L} = \mathbb{E}_{(x, y, a) \sim p(x, y, a)} \left[ \ell(y, f(x)) \right].
\end{equation}
Given the set of parameters $\theta$ of our model $f $, the gradient of $\mathcal{L}_{ext}$ with respect to $ \theta$ can be derived from \cref{eq:relation_losses}: 
\begin{equation}
\label{eq:grads}
    \nabla_{\theta} \mathcal{L}_{ext} = \nabla_{\theta} \mathcal{L}_{lb} + \beta \cdot\left( \nabla_{\theta} \mathcal{L}_{lb} - \nabla_{\theta} \mathcal{L}_b \right).
\end{equation}
Our target gradient vector $\nabla_{\theta} \mathcal{L}_{ext}$ in \cref{eq:grads} is a linear extrapolation of the two gradient vectors $\nabla_{\theta} \mathcal{L}_{lb}$ and $\nabla_{\theta} \mathcal{L}_{b}$, and accordingly, we refer to $\beta$ as the extrapolation factor.
Because the less biased batch has a less skewed conditional attribute distribution compared to the biased batch (as shown in \cref{eq:inequality_p_a_y}), extrapolating their gradients and toward the less biased gradient forms a new gradient ($\mathcal{L}_{ext}$) that leads to learning even more debiased representation for some values of the extrapolation factor $\beta>0$. A visual representation of extrapolation is shown in \cref{fig:subfig3}.
\subsubsection{GERNE as a general framework for debiasing}
\label{sec:general_framework}
Minimizing our target loss $\mathcal{L}_{ext}$ simulates minimizing the loss of class-balanced batches with the following conditional distribution of $(y, a) \in \mathcal{G}$ :
\begin{equation}
\label{eq:p_ext}
p_{ext}(a|y) = \alpha_{ya} + c \cdot(\beta+1) \cdot\left( \frac{1}{A} - \alpha_{ya} \right).
\end{equation}
We provide the full proof in \cref{appendix:proof_p_ext}.

Based on  \cref{eq:p_ext}, we can establish the link between GERNE and other methods for different values of $\beta, c$: 
\begin{itemize}
    \item[-] For $\beta = -1$, $\mathcal{L}_{ext} = \mathcal{L}_b$ and GERNE is equivalent to class-balanced ERM method.
   \item[-] For $c=1$ and $\beta=0$, $p_{ext}(a|y)=\frac{1}{A}$, and GERNE matches Resampling ~\cite{idrissi2022simple}, which samples equally from all groups ($B_{rs}$ in \cref{fig:subfig2}, with gradient of the loss computed on it denoted as $\nabla_{\theta} \mathcal{L}_{rs}$ in \cref{fig:subfig3}). 
    \item[-]  For $c \cdot (\beta+1) =1$, we also have $p_{ext}(a|y)=\frac{1}{A}$, and $\mathcal{L}_{ext}$ is, in expectation, equivalent to $\mathcal{L}_{rs}$. However, their loss variances differ. In fact, GERNE permits controlling the variance of its loss through its hyperparameters ($c, \beta$),  which may help escape sharp minima ~\cite{escape_r} and improve generalization ~\cite{keskar2017on}. The derivation of the variance of GERNE's loss is detailed in \cref{appendix:GERNE_variance}.
    \item[-]  For $c \cdot (\beta+1) >1$,  $p_{ext}(a|y) >\frac{1}{A} \text{ if } \alpha_{ya}<\frac{1}{A}$ (also $p_{ext}(a|y) < \frac{1}{A} \text{ if } \alpha_{ya}>\frac{1}{A}$). In this case, GERNE simulates batches where the underrepresented groups (i.e., those with $\alpha_{ya}<\frac{1}{A}$) are oversampled.
\end{itemize}
\subsubsection{\texorpdfstring{Upper and lower bounds of $\beta$}{Upper and lower bounds of beta}}
Having $ p_{ext}(a|y)$ in \cref{eq:p_ext} within $[0,1]$, $\beta$ should satisfy:
\begin{equation}
\begin{split}
\label{eq:intervals_beta}
\max_{\substack{{(y,a) \in \mathcal{G}} \\ \alpha_{ya} \neq \frac{1}{A}}} \min(i_{ya}^1, i_{ya}^2) \leq \beta \leq \min_{\substack{{(y,a) \in \mathcal{G}} \\ \alpha_{ya} \neq \frac{1}{A}}}\max(i_{ya}^1,i_{ya}^2),
\end{split}
\end{equation}
where:
$ i_{ya}^1 = -\frac{\alpha_{ya}}{c\cdot (\frac{1}{A}-\alpha_{ya})} -1, i_{ya}^2 = \frac{1-\alpha_{ya}}{c\cdot(\frac{1}{A}-\alpha_{ya})} -1 $.\\
These bounds are used when tuning $\beta$. In \cref{appendix:simplified_bounds}, we simplify these bounds to $[\beta_{\min},\beta_{\max}]=[-1, i_{y''a''}^1]$, where $(y'',a'') = \arg \max_{(y,a) \in \mathcal{G}} \alpha_{ya}$. Note that $\beta$ doesn't affect $p_{ext}(a|y)$ for $\alpha_{ya}=\frac{1}{A}$ according to \cref{eq:p_ext}.
\subsubsection{\texorpdfstring{Tuning $\beta$ to minimize the risk for worst-case group}{Tuning beta to minimize the risk for worst-case group}}
\label{sec:beta_wga}
\cref{eq:relation_losses} can be rewritten as follows (detailed in \cref{appendix:proof_p_ext}):
\begin{equation}
\label{eq:ld_lg}
    \mathcal{L}_{ext} = \frac{1}{K}\cdot\sum_{g=(y,a) \in \mathcal{G}}  p_{ext}(a|y) (\beta) \cdot L_g,
\end{equation}
where 
\begin{equation}
\label{eq:lg}
L_g = \mathbb{E}_{x \sim p(x| (y, a)=g)}\left[ \ell(y, f(x)) \right].
\end{equation}
In the presence of spurious correlations, minority or less-represented groups often experience higher risks, primarily due to the model’s limited exposure to these groups during training ~\cite{johnson2020effects}. Taking this into consideration, we define $g'=(y',a') = \arg \min_{(y,a) \in \mathcal{G}} \alpha_{ya}$. 
Since $L_{g'}$ is weighted by $p_{ext}(a'|y') $, increasing $\beta$ beyond $\frac{1}{c}-1$ assigns more weight to $L_{g'}$ in \cref{eq:ld_lg} than any other group loss (all groups' losses are equally weighted when $c\cdot (\beta+1) = 1$). This increase in $\beta$ encourages the model to prioritize reducing the loss of the underrepresented group $g'$ during training, therefore minimizing the risk for the worst-case group. 

We outline the detailed steps of our approach for the known attribute case in \cref{alg:known_attr_algo}.
\label{appendix:algo1.}
\begin{algorithm}
\caption{GERNE for the known attribute case}
\label{alg:known_attr_algo}
\textbf{Input:} $\mathcal{X}_{y, a} \subseteq \mathcal{X}$ for $ y \in \mathcal{Y} $ and $a \in \mathcal{A}$, $f$ with initial $\theta = \theta_0 $, $\#$ epochs $E$, batch size per label $B$, $\#$ classes $K$, $\#$ attributes $A$, learning rate $\eta$.
\begin{algorithmic}[1]
\State Choose $c\in(0, 1]$ and $\beta\in[\beta_{\min},\beta_{\max}]$ via grid search.
\For {epoch $= 1$ to $E$}
    \State Biased Batch $ B_b = \emptyset$, Less Biased Batch $ B_{lb} = \emptyset$
    \For {$ (y,a) \in \mathcal{G}$}
    \State Sample a mini-batch $ B_b^{y,a} = \{(x, y)\} \subseteq \mathcal{X}_{y,a}$ of size $\alpha_{y, a}\cdot B$;
    \State $B_b = B_b \cup  B_b^{y, a}$
    \State Sample a mini-batch $ B_{lb}^{y,a} = \{(x, y)\}\subseteq \mathcal{X}_{y,a} $ of size $((1-c)\cdot\alpha_{y, a}+\frac{c}{A}) \cdot B$
    \State $B_{lb} = B_{lb} \cup  B_{lb}^{y, a}$
    \EndFor
    \State Compute $\mathcal{L}_b, \mathcal{L}_{lb}$ on $B_b, B_{lb}$, respectively. Then, compute $\nabla_{\theta} \mathcal{L}_{b}$ and $\nabla_{\theta} \mathcal{L}_{lb}$.
    \State Compute $\nabla_{\theta} \mathcal{L}_{ext}$  using \cref{eq:grads}.
    \State Update parameters (SGD): $\theta \leftarrow \theta - \eta \cdot \nabla_\theta \mathcal{L}_{ext}$
\EndFor
\end{algorithmic}
\end{algorithm}
\subsection{GERNE for  the Unknown Attributes Case}
\label{sec:unknown_attr}
If the attributes are unavailable during training, it is not possible to directly sample less biased batches. To address this, we follow the previous work ~\cite{LfF, JTT, zhang_correct-n-contrast_2022} by training a standard ERM model $\tilde{f}$ and using its predictions to create pseudo-attributes $\tilde{a}$. Since $\tilde{f}$ is trained on biased batches, it tends to rely on spurious correlations, resulting in biased predictions. Leveraging these predictions, we classify samples into easy---those with high-confidence predictions, where the spurious correlations likely hold--- and hard --- those with low-confidence predictions, where the spurious correlations may not hold.
After training $\tilde{f}$, we select a threshold $t \in (0,1)$ and construct pseudo-attributes based on model predictions as follows:
For each class $y$, we compute the predictions $\tilde{y}_i = p(y|x_i) = \text{softmax}(\tilde{f}(x_i))_y$ for each $x_i \in \mathcal{X}_y$. We then split them into two non-empty subsets: The first subset contains the smallest \( \lfloor t \cdot |\mathcal{X}_y| \rfloor \) values, and the corresponding samples form the group $\mathcal{X}_{y, \tilde{a}=1}$. The remaining samples forms the group $\mathcal{X}_{y, \tilde{a}=2}$. This process ensures that each set $\mathcal{X}_y$ is divided into two disjoint and non-empty groups. Consequently, the pseudo-attribute space consists of two values, denoted as $\mathcal{\tilde{A}} = \{1, 2\}$ (i.e., $\tilde{A} =2$) with $\tilde{\mathcal{G}} = \mathcal{Y} \times \mathcal{\tilde{A}}$ replacing $\mathcal{G}$ in the unknown attribute case. $t$ is a hyperparameter, and we outline the detailed steps of GERNE for the unknown case in \cref{appendix:algorithm2}.
\subsubsection{\texorpdfstring{Tuning $\beta$ to control the unknown conditional distribution of an attribute $a$ in class $y$}{Tuning beta to control the unknown conditional distribution of an attribute a in class y}}
After creating the pseudo-attributes and defining the pseudo-groups, we consider forming a new batch of size $B$ by uniformly sampling $\gamma \cdot B$ examples from group $\mathcal{X}_{y,\tilde{a}=1}$ and $(1-\gamma) \cdot B$ examples from group $\mathcal{X}_{y, \tilde{a}=2}$, where $\gamma \in [0,1], \gamma \cdot B \in \mathbb{N}$.
The resulting conditional distribution of an attribute $a$ given $y$ in the constructed batch is:
\begin{align}
p_{B}(a|y) &= \sum_{\tilde{a} \in \mathcal{\tilde{A}}} p_{B}(\tilde{a}|y) \cdot p(a|\tilde{a}, y).
\end{align}
Because the max/min value of a linear program must occur at a vertex, we have for $p(a|\tilde{a},y)=p_{\tilde{a},y}(a)$:
\begin{equation}
\label{eq:impossible_cond}
\forall \gamma \in [0,1], \min_{\tilde{a} \in \mathcal{\tilde{A}}}p_{\tilde{a},y}(a) \leq  p_{B}(a|y)    \leq \max_{\tilde{a} \in \mathcal{\tilde{A}}}p_{\tilde{a},y}(a).
\end{equation}
This means that if: 
$\max_{\tilde{a}}p_{\tilde{a},y}(a) < \frac{1}{A}$$( \min_{\tilde{a}}p_{\tilde{a},y}(a) > \frac{1}{A})$, then there is no value for $\gamma$ can yield a batch with $p_{B}(a|y) > \frac{1}{A}$$(p_{B}(a|y)<\frac{1}{A})$ via sampling from the pseudo-groups. 
\paragraph{Proposition 1.} In case of unknown attributes, GERNE can simulate creating batches with more controllable conditional attribute distribution (i.e., $p_{B}(a|y) > \max_{\tilde{a} \in \mathcal{\tilde{A}}}p_{\tilde{a},y}(a) \text{ or } p_{B}(a|y) < \min_{\tilde{a} \in \mathcal{\tilde{A}}}p_{\tilde{a},y}(a)$). We provide the proof of this proposition in \cref{appendix:useful_beta}.
\section{Experiments}
\label{sec:experiments}
To evaluate the general applicability of GERNE, we assess its performance across five computer vision and one natural language processing benchmarks: Colored MNIST (C-MNIST) ~\cite{arjovsky2020invariantriskminimization, lee_learning_2021}, Corrupted CIFAR-10 (C-CIFAR-10)~\cite{hendrycks2019benchmarking, LfF}, Biased FFHQ (bFFHQ) ~\cite{karras2019style, lee_learning_2021}, Waterbird ~\cite{waterbirds}, CelebA ~\cite{celeba}, and CivilComments ~\cite{civilcomments}.
We categorize these datasets into two groups: Datasets-1 and Datasets-2. Datasets-1 comprises the first three datasets mentioned above and is used to evaluate GERNE’s performance without data augmentation. Datasets-2 consists of the remaining three datasets, for which we follow the experimental setup described in ~\cite{ChangeIsHard} to ensure a fair comparison. 
\subsection{Experiments on Datasets-1}
\paragraph{Datasets.}
C-MNIST is an extension of the MNIST dataset ~\cite{mnist} where each digit class is predominantly associated with a specific color. This introduces a spurious correlation between the digit label (target) and color (attribute).
C-CIFAR-10 modifies CIFAR-10 by applying specific texture patterns to each object class ~\cite{hendrycks2019benchmarking}, making texture a spurious feature. Both C-MNIST and C-CIFAR-10 include versions with varying degrees of spurious correlation, reflected by the minority group ratios of 0.5\%, 1\%, 2\%, and 5\% in the training and validation sets.
The bFFHQ dataset comprises human face images, with ``age'' and ``gender'' as the target and spurious attributes, respectively. The majority of female faces are young, while the majority of males are old. The minority group ratio in the training set is 0.5\%.
\paragraph{Evaluation metrics.} We follow the evaluation protocols of prior work \cite{LfF, liu_avoiding_2023, lee_learning_2021}. For C-MNIST and C-CIFAR-10, we report Group-Balanced Accuracy (GBA) on the test set. For bFFHQ, we evaluate performance based on the accuracy of the minority group.
\paragraph{Baselines.} For the known attribute case, we compare GERNE with Group DRO ~\cite{DRO} and Resampling ~\cite{idrissi2022simple}. For the unknown attribute case, we consider ERM ~\cite{ERM}, JTT ~\cite{JTT}, LfF ~\cite{LfF}, DFA ~\cite{lee_learning_2021}, LC ~\cite{liu_avoiding_2023}, and DeNetDM ~\cite{vadakkeeveetil_denetdm_2024}. 
\paragraph{Implementation details.} We adopt the same model architectures as the baselines and use SGD optimizer across all three datasets. More details are provided in  \cref{appendix:baselines_impdetails}.
\setlength{\tabcolsep}{4pt}
\begin{table*}[t]
\centering
\caption{Performance comparison of GERNE and baselines on the C-MNIST, C-CIFAR-10, and bFFHQ datasets. We report GBA (\%) with standard deviation over three trials for C-MNIST and C-CIFAR-10 across varying minority ratios, and minority group accuracy (\%) for bFFHQ. DeNetDM results are from \cite{vadakkeeveetil_denetdm_2024}, and Resampling results are generated using GERNE with $c=1, \beta=0$. All other results are from ~\cite{liu_avoiding_2023}. $\checkmark/ \times$ indicates known/unknown training attributes. The \textbf{best} results are marked in bold, and the \underline{second-best} are underlined.}

\begin{adjustbox}{width=1\textwidth}
\scriptsize
\begin{tabular}{lcccccccccc}
\toprule
\multirow{2}{*}{\raisebox{-1.5ex}{\textbf{Methods}}} & \multirow{2}{*}{\raisebox{0.5ex}{\textbf{Group}}} & \multicolumn{4}{c}{\textbf{C-MNIST}} & \multicolumn{4}{c}{\textbf{C-CIFAR-10}} & \multirow{2}{*}{\raisebox{-1.5ex}{\textbf{bFFHQ}}} \\
\cmidrule(lr){3-6} \cmidrule(lr){7-10}
 & {\raisebox{-0.5ex}{\textbf{Info}}} & 0.5 & 1 & 2 & 5 & 0.5 & 1 & 2 & 5 \\
\midrule
Group DRO & \checkmark & 63.12 & 68.78 & 76.30 & 84.20 & 33.44 & 38.30 & 45.81 & 57.32 & -  \\
Resampling & \checkmark & 77.68 \scriptsize{$\pm$0.89} & 84.36 \scriptsize{$\pm$0.21}& 88.15 \scriptsize{$\pm$0.11} & 91.98 \scriptsize{$\pm$0.08} & 45.10 \scriptsize{$\pm$0.60} & 50.08 \scriptsize{$\pm$0.42} & 54.85 \scriptsize{$\pm$0.30} & 62.16 \scriptsize{$\pm$0.05} &  72.13 \scriptsize{$\pm$0.90} \\
GERNE (ours)& \checkmark & \textbf{77.79 \scriptsize{$\pm$0.90}} & \textbf{84.47 \scriptsize{$\pm$0.37}} & \textbf{88.30 \scriptsize{$\pm$0.20}} & \textbf{92.16 \scriptsize{$\pm$0.10}} & \textbf{45.34 \scriptsize{$\pm$0.60}} &  \textbf{50.84 \scriptsize{$\pm$0.17}}  &\textbf{55.51 \scriptsize{$\pm$0.10}} & \textbf{62.40 \scriptsize{$\pm$0.27}} & \textbf{85.20 \scriptsize{$\pm$0.86}} \\
\midrule
ERM & $\times$ & 35.19 \scriptsize{$\pm$3.49} & 52.09 \scriptsize{$\pm$ 2.88} & 65.86 \scriptsize{$\pm$ 3.59} & 82.17 \scriptsize{$\pm$ 0.74} & 23.08 \scriptsize{$\pm$ 1.25} & 28.52 \scriptsize{$\pm$ 0.33} & 30.06 \scriptsize{$\pm$ 0.71} & 39.42 \scriptsize{$\pm$ 0.64} & 56.70 \scriptsize{$\pm$ 2.70} \\ 
JTT & $\times$ & 53.03 \scriptsize{$\pm$ 3.89} & 62.90 \scriptsize{$\pm$ 3.01} & 74.23 \scriptsize{$\pm$ 3.21} & 84.03 \scriptsize{$\pm$ 1.10} & 24.73 \scriptsize{$\pm$ 0.60} & 26.90 \scriptsize{$\pm$ 0.31} & 33.40 \scriptsize{$\pm$ 1.06} & 42.20 \scriptsize{$\pm$ 0.31} & 65.30 \scriptsize{$\pm$ 2.50} \\
LfF & $\times$ & 52.50 \scriptsize{$\pm$ 2.43} & 61.89 \scriptsize{$\pm$ 4.97} & 71.03 \scriptsize{$\pm$ 1.14} & 84.79 \scriptsize{$\pm$ 1.09} & 28.57 \scriptsize{$\pm$ 1.30} & 33.07 \scriptsize{$\pm$ 0.77} & 39.91 \scriptsize{$\pm$ 1.30} & 50.27 \scriptsize{$\pm$ 1.56} & 62.20 \scriptsize{$\pm$ 1.60} \\
DFA & $\times$ & 65.22 \scriptsize{$\pm$ 4.41} & 81.73 \scriptsize{$\pm$ 2.34} & 84.79 \scriptsize{$\pm$ 0.95} & 89.66 \scriptsize{$\pm$ 1.09} & 29.75 \scriptsize{$\pm$ 0.71} & 36.49 \scriptsize{$\pm$ 1.79} & 41.78 \scriptsize{$\pm$ 2.29} & 51.13 \scriptsize{$\pm$ 1.28} & 63.90 \scriptsize{$\pm$ 0.30} \\
LC & $\times$ & \underline{71.25 \scriptsize{$\pm$ 3.17}} & \underline{82.25 \scriptsize{$\pm$ 2.11}} & \underline{86.21 \scriptsize{$\pm$ 1.02}} & \textbf{91.16 \scriptsize{$\pm$ 0.97}} & 34.56 \scriptsize{$\pm$ 0.69} & 37.34 \scriptsize{$\pm$ 1.26} & 47.81 \scriptsize{$\pm$ 2.00} & 54.55 \scriptsize{$\pm$ 1.26} & 69.67 \scriptsize{$\pm$ 1.40} \\
DeNetDM & $\times$ &- & - & - & - & \underline{38.93 \scriptsize{$\pm$ 1.16}} & \underline{44.20 \scriptsize{$\pm$ 0.77}} & \underline{47.35 \scriptsize{$\pm$ 0.70}} & \underline{56.30 \scriptsize{$\pm$ 0.42}} & \underline{75.70 \scriptsize{$\pm$ 2.80}} \\
GERNE (ours) & $\times$ & \textbf{77.25 \scriptsize{$\pm$ 0.17}} &\textbf{83.98 \scriptsize{$\pm$ 0.26}} & \textbf{87.41 \scriptsize{$\pm$ 0.31}} & \underline{90.98 \scriptsize{$\pm$ 0.13}} & \textbf{39.90\scriptsize{$\pm$ 0.48}} &  \textbf{45.60\scriptsize{$\pm$ 0.23}} &   \textbf{50.19\scriptsize{$\pm$ 0.18}} & \textbf{56.53 \scriptsize{$\pm$ 0.32}} & \textbf{76.80 \scriptsize{$\pm$ 1.21}} \\
\bottomrule
\end{tabular}
\end{adjustbox}
\label{table:synth_data}
\end{table*}
\paragraph{Results.} 
\cref{table:synth_data} compares GERNE with baselines for both known and unknown attribute cases. All baseline results are adopted from ~\cite{liu_avoiding_2023}, except DeNetDM, which is sourced from ~\cite{vadakkeeveetil_denetdm_2024}. When the attributes are known, GERNE outperforms Group DRO by a significant margin on C-MNIST and C-CIFAR-10 datasets. The improvement in performance ranges from about 5\% on C-CIFAR-10 with 5\% of minority group and up to 16\% on C-MNIST with 1\% of minority group. Furthermore, GERNE outperforms Resampling ~\cite{idrissi2022simple} by over 13\% on bFFHQ and consistently surpasses it across all other ratios. 
Our explanation behind GERNE's superior performance over Resampling is that the latter tends to present the majority and minority groups equally in the sampled batches during training, and the model $f$ tends to prioritize learning the easy-to-learn spurious features associated with the majority group (e.g., the colors in C-MNIST), leading to learning biased representation and poorer generalization. In contrast, GERNE undermines learning the spurious features by directing the learning process more in the debiasing direction, thanks to the extrapolation factor. 
For the unknown attribute case, GERNE outperforms all baselines, except on C‑MNIST with 5\% of minority group (ranks second), while maintaining a lower standard deviation.
At this 5\% minority ratio, LC achieves slightly higher accuracy---likely benefiting from its use of data augmentation to increase the diversity of the samples in the minority group. We exclude DeNetDM’s results on C-MNIST, as the authors use a different version of this dataset.
\subsection{Experiments on Datasets-2}
\paragraph{Datasets.}  Waterbirds ~\cite{waterbirds}  contains bird images with spurious correlations between bird type and background: Most waterbirds appear with water backgrounds, while most landbirds appear with land backgrounds. CelebA ~\cite{celeba} involves classifying hair color (blond, non-blond), with gender (male, female) as the spurious attribute: Most blond images depict females. CivilComments ~\cite{civilcomments} is a binary toxic comment classification dataset, where the spurious attribute marks references to eight different demographic identities (male, female, LGBTQ, Christian, Muslim, other religions, Black, and White).
\paragraph{Evaluation metrics.} We follow the same evaluation strategy from ~\cite{ChangeIsHard} for model selection and hyperparameter tuning.
When attributes are known in both training and validation, we use the worst-group test accuracy as the evaluation metric.
When attributes are unknown in training, but known in validation, we use the worst-group validation accuracy. When attributes are unavailable in both, we use the worst-class validation accuracy.
\paragraph{Baselines.}  For each dataset, we select the best three methods reported in ~\cite{ChangeIsHard}. We end up with ERM~\cite{ERM}, Group DRO~\cite{DRO}, DFR ~\cite{DFR}, LISA ~\cite{LISA}, ReSample ~\cite{ReWeightResample}, Mixup ~\cite{Mixup}, ReWeightCRT ~\cite{ReWeightCRT}, ReWeight ~\cite{ReWeightResample}, CBLoss ~\cite{CBLoss}, BSoftmax~\cite{BSoftmax} and SqrtReWeight ~\cite{ChangeIsHard}. We also report the results for CnC ~\cite{zhang_correct-n-contrast_2022} as it adopts similar training settings. 
\paragraph{Implementation details.}
We employ the same data augmentation techniques, optimizers and pretrained models described in ~\cite{ChangeIsHard}. Further details are in \cref{appendix:baselines_impdetails_dataset2}. 
\paragraph{Results}
\cref{table:real_data} shows the worst-group accuracy (WGA) of the test set for GERNE compared to the baseline methods under the evaluation strategy explained above. In the known attributes case, GERNE achieves the highest accuracy on CelebA and CivilComments, and ranks second on Waterbirds, following DFR. 
In case of unknown attributes in the training set but known in validation, our approach again attains the best results on Waterbirds and CivilComments datasets and remains competitive on CelebA, closely following the top two baselines' results. In particular, DFR uses the validation set to train the model, whereas GERNE employs it only for model selection and hyperparameter tuning. We include a comparison between DFR and GERNE when using the validation set for training in \cref{appendix:GERNE_limited_info}.
When attributes are unknown in both the training and validation sets, GERNE achieves the best results on Waterbirds and CelebA. However, we observe a significant drop in accuracy on CelebA compared to the second case (known attributes only in validation), while this drop is less pronounced on Waterbirds.  The difference can be attributed to the use of worst-class accuracy as the evaluation metric. In CelebA's validation set, the majority of blond hair images exhibit spurious correlations (female images), leading the model selection process to favor the majority group while disregarding the minority group. In contrast, the validation set in Waterbirds is group-balanced within each class, leading to only a slight decrease in performance between the second and third case. This highlights the critical role of having access to the attributes in the validation set ---or at least a group-balanced validation set--- for model selection when using GERNE.
\setlength{\tabcolsep}{4pt}
\begin{table}[ht]
\centering
\caption{Performance comparison of GERNE and baseline methods on Waterbirds, CelebA, and CivilComments. We report the worst-group test accuracy (\%) and standard deviation over three trials for each dataset. Baseline results are sourced from ~\cite{ChangeIsHard} as the same experimental settings are adopted. $\checkmark / \checkmark$ denotes known attributes in training and validation sets. $\times / \checkmark$ indicates attributes are known only in the validation set, while $\times / \times$ signifies that attributes are unknown in both sets. \textbf{Best} results are highlighted in bold, and the \underline{second-best} are underlined.}
\begin{adjustbox}{width=0.47\textwidth}
\scriptsize 
\begin{tabular}{lcccc}
\toprule
\multirow{2}{*}{\raisebox{-0.7ex}{\textbf{Methods}}}  & {\textbf{Group Info}} & \multirow{2}{*}{\raisebox{-0.7ex}{\textbf{Waterbirds}}} & \multirow{2}{*}{\raisebox{-0.7ex}{\textbf{CelebA}}}&  \multirow{2}{*}{\raisebox{-0.7ex}{\textbf{Civil-}}} \\
\cmidrule(lr){2-2}
 & \textbf{train}/\textbf{val attr. } & & & \textbf{Comments}\\
\midrule
ERM &  \checkmark/\checkmark & 69.10 \scriptsize{$\pm$ 4.70} & 62.60 \scriptsize{$\pm$ 1.50} & 63.70 \scriptsize{$\pm$ 1.50} \\ 
Group DRO & \checkmark/\checkmark & 78.60 \scriptsize{$\pm$ 1.00} & 89.00 \scriptsize{$\pm$ 0.70} & 70.60 \scriptsize{$\pm$ 1.20}   \\
ReWeight & \checkmark/\checkmark & 86.90 \scriptsize{$\pm$ 0.70} & 89.70 \scriptsize{$\pm$ 0.20}  & 65.30 \scriptsize{$\pm$ 2.50} \\
ReSample & \checkmark/\checkmark & 77.70 \scriptsize{$\pm$ 1.20}& 87.40 \scriptsize{$\pm$ 0.80}& 73.30 \scriptsize{$\pm$ 0.50} \\
CBLoss & \checkmark/\checkmark & 86.20 \scriptsize{$\pm$ 0.30} & 89.40 \scriptsize{$\pm$ 0.70}  & 73.30 \scriptsize{$\pm$ 0.20} \\
DFR & \checkmark/\checkmark & \textbf{91.00 \scriptsize{$\pm$ 0.30}} & \underline{90.40 \scriptsize{$\pm$ 0.10}} & 69.60 \scriptsize{$\pm$ 0.20} \\
LISA & \checkmark/\checkmark & 88.70 \scriptsize{$\pm$ 0.60} & 86.50 \scriptsize{$\pm$ 1.20} & \underline{73.70 \scriptsize{$\pm$ 0.30}} \\
GERNE (ours) & \checkmark/\checkmark & \underline{90.20 \scriptsize{$\pm$ 0.22}} & \textbf{91.98 \scriptsize{$\pm$ 0.15}} & \textbf{74.65 \scriptsize{$\pm$ 0.20}}\\
\midrule
ERM & $\times$/\checkmark & 69.10 \scriptsize{$\pm$ 4.70} & 57.60 \scriptsize{$\pm$ 0.80}& 63.20 \scriptsize{$\pm$ 1.20} \\ 
Group DRO & $\times$/\checkmark & 73.10 \scriptsize{$\pm$ 0.40} & 78.50 \scriptsize{$\pm$ 1.10} & 69.50 \scriptsize{$\pm$ 0.70}   \\
ReWeight & $\times$/\checkmark & 72.50 \scriptsize{$\pm$ 0.30} & 81.50 \scriptsize{$\pm$ 0.90} & \underline{69.90 \scriptsize{$\pm$ 0.60}} \\
DFR & $\times$/\checkmark & \underline{89.00 \scriptsize{$\pm$ 0.20}} &\underline{86.30 \scriptsize{$\pm$ 0.30}}& 63.90 \scriptsize{$\pm$ 0.30} \\
Mixup & $\times$/\checkmark & 78.20 \scriptsize{$\pm$ 0.40} & 57.80 \scriptsize{$\pm$ 0.80}& 66.10 \scriptsize{$\pm$ 1.30} \\
LISA & $\times$/\checkmark & 78.20 \scriptsize{$\pm$ 0.40} & 57.80 \scriptsize{$\pm$ 0.80} & 66.10 \scriptsize{$\pm$ 1.30} \\
BSoftmax & $\times$/\checkmark & 74.10 \scriptsize{$\pm$ 0.90} & 83.30 \scriptsize{$\pm$ 0.30}& 69.40 \scriptsize{$\pm$ 1.20} \\
ReSample & $\times$/\checkmark & 70.00 \scriptsize{$\pm$ 1.00} & 82.20 \scriptsize{$\pm$ 1.20} & 68.20 \scriptsize{$\pm$ 0.70} \\
CnC & $\times$/\checkmark & 88.50\scriptsize{$\pm$ 0.30} & \textbf{88.80\scriptsize{$\pm$ 0.90}} & 68.90\scriptsize{$\pm$ 2.10} \\
GERNE (ours) & $\times$/\checkmark & \textbf{90.21 \scriptsize{$\pm$ 0.42}} &  86.28 \scriptsize{$\pm$ 0.12} & \textbf{71.00 \scriptsize{$\pm$ 0.33}} \\

\midrule
ERM & $\times$/$\times$ & 69.10 \scriptsize{$\pm$ 4.70} & 57.60 \scriptsize{$\pm$ 0.80}  & 63.20 \scriptsize{$\pm$ 1.20} \\ 
Group DRO & $\times$/$\times$ & 73.10 \scriptsize{$\pm$ 0.40} & 68.30 \scriptsize{$\pm$ 0.90} & 61.50 \scriptsize{$\pm$ 1.80}   \\
DFR & $\times$/$\times$ & \underline{89.00 \scriptsize{$\pm$ 0.20}} & 73.70 \scriptsize{$\pm$ 0.80} & 64.40 \scriptsize{$\pm$ 0.10} \\
Mixup & $\times$/$\times$ & 77.50 \scriptsize{$\pm$ 0.70} & 57.80 \scriptsize{$\pm$ 0.80} & 65.80 \scriptsize{$\pm$ 1.50} \\
LISA & $\times$/$\times$ & 77.50 \scriptsize{$\pm$ 0.70} & 57.80 \scriptsize{$\pm$ 0.80} & 65.80 \scriptsize{$\pm$ 1.50} \\
ReSample & $\times$/$\times$ & 70.00 \scriptsize{$\pm$ 1.00} & \underline{74.10 \scriptsize{$\pm$ 2.20}} & 61.00 \scriptsize{$\pm$ 0.60} \\
ReWeightCRT & $\times$/$\times$ & 76.30 \scriptsize{$\pm$ 0.20} & 70.70 \scriptsize{$\pm$ 0.60}& 64.70 \scriptsize{$\pm$ 0.20} \\
SqrtReWeight & $\times$/$\times$ & 71.00 \scriptsize{$\pm$ 1.40} & 66.90 \scriptsize{$\pm$ 2.20}& \textbf{68.60 \scriptsize{$\pm$ 1.10}} \\
CRT & $\times$/$\times$ & 76.30 \scriptsize{$\pm$ 0.80} & 69.60 \scriptsize{$\pm$ 0.70} &\underline{67.80 \scriptsize{$\pm$ 0.30}} \\
GERNE (ours) & $\times$/$\times$ & \textbf{89.88 \scriptsize{$\pm$ 0.67}} & \textbf{74.24 \scriptsize{$\pm$ 2.51}} &  63.10 \scriptsize{$\pm$ 0.22}  \\
\bottomrule
\end{tabular}
\end{adjustbox}
\label{table:real_data}
\end{table}

\subsection{GERNE vs. Balancing Techniques}
\label{sec:balance}
Balancing techniques have been shown to achieve state-of-the-art results, while remaining easy to implement ~\cite{idrissi2022simple, ChangeIsHard}.
While Resampling often outperforms Reweighting when combined with stochastic gradient algorithms~\cite{resampling_r}, we show in \cref{table:synth_data} that GERNE consistently outperforms Resampling in both group-balanced accuracy (GBA) and minority group accuracy. 
This highlights the flexibility of GERNE to adapt to maximize both metrics, and its superior performance in comparison to Resampling and other balancing techniques, as further supported by the results in \cref{table:real_data}. In \cref{appendix:GERNE_variance}, we provide a detailed ablation study comparing GERNE to an equivalent ``sampling+weighting'' approach with matching loss expectation, and demonstrate how GERNE can leverage its controllable loss variance (by the hyperparameters $c, \beta$) to escape sharp minima.
\subsection{Ablation Study}
\begin{figure}[htp]
    \centering
    \includegraphics[width=0.47\textwidth]{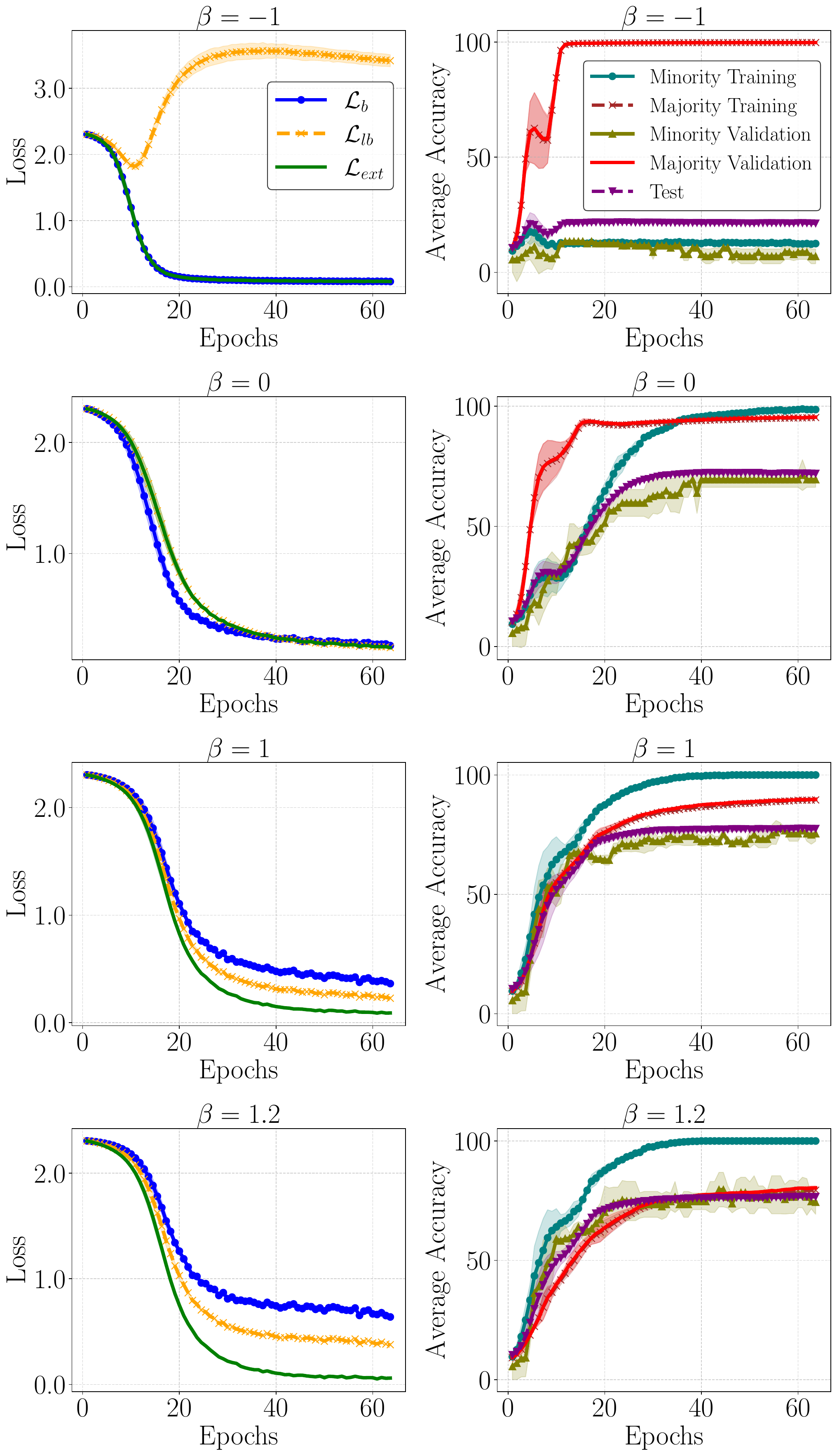}
    \caption{The impact of tuning $\beta \in \{-1,0,1,1.2\}$  on debiasing the model.  On the left column, we plot the training losses  $\mathcal{L}_b, \mathcal{L}_{lb}$ and the target loss $\mathcal{L}_{ext}$. On the right column, we plot the average accuracy of the minority and majority groups in both training and validation sets, as well as the average accuracy of the unbiased test set. Each plot represents the mean and standard deviation calculated over three runs with different random seeds.}
    \label{fig:tuning_beta}
\end{figure}

\begin{figure}[t]
    \centering    \includegraphics[width=0.47\textwidth]{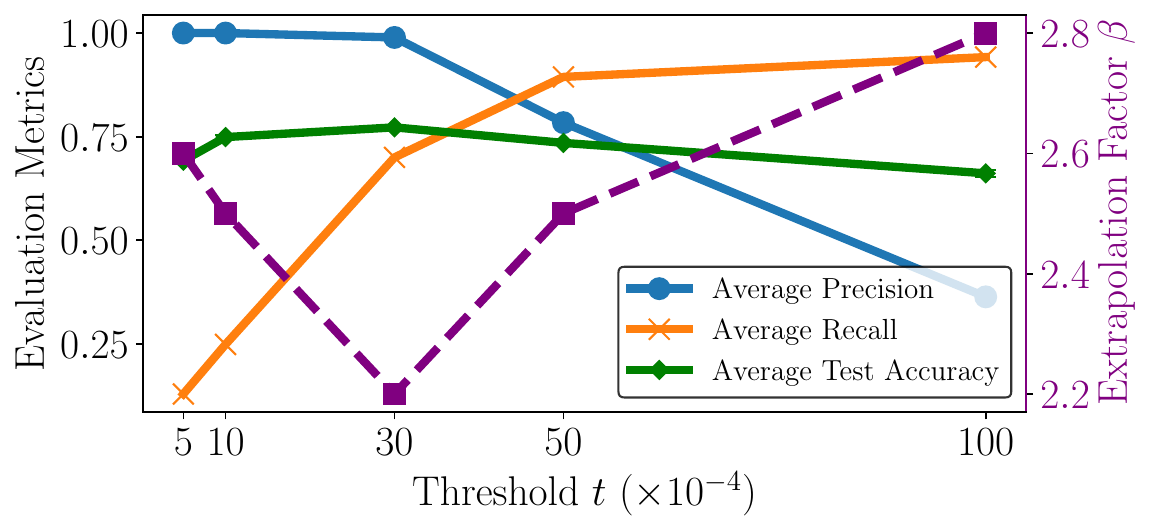}
        \caption{The effect of the threshold \(t\) used to generate pseudo-attributes on the extrapolation factor \(\beta\) and model performance. We plot the average precision and recall over pseudo-minority groups \((y, \tilde{a}=1)\), averaged across all classes \(y\). For each \((y, \tilde{a}=1)\), precision is defined as the fraction of minority samples among all samples in that group, and recall is the fraction of those minority samples relative to all minority samples in class \(y\). We also report the best achievable test accuracy, along with the corresponding extrapolation factor \(\beta\), across different threshold values.}
        \label{fig:threshold}
\end{figure}

\paragraph{Tuning the extrapolation factor $\beta$.}
The value of $\beta$ in \cref{eq:grads} plays a critical role in guiding the model toward learning debiased representation (i.e., reducing reliance on spurious features and improving generalization).  In \cref{fig:tuning_beta}, we illustrate the effect of tuning $\beta$ on the learning process using C-MNIST with $0.5\%$ of minority group in the known attributes case. We show results for $\beta \in \{-1,0,1,1.2\} \text{ with } c= 0.5$. For $\beta=-1$, our target loss $\mathcal{L}_{ext}$ in \cref{eq:relation_losses} equals the biased loss $\mathcal{L}_b$, which leads to learning a biased model that exhibits high accuracy on the majority group, yet demonstrates poor performance on both the minority group and the unbiased test set. 
As $\beta$ increases (e.g., $\beta=0, \beta=1$), the model starts learning more intrinsic features. This is evident from the improved performance on the minority group in the validation set, as well as on the unbiased test set.
However, as the extrapolation factor $\beta$ continues to increase, the model begins to exhibit higher variance during the training process, as shown for $\beta=1.2$, ultimately leading to divergence when $\beta$ exceeds the upper bound defined in \cref{eq:intervals_beta} ($1.22$ in this case). While GERNE appears to be sensitive to small variations in $\beta$ (e.g., 1.2 to 1.22), we show in \cref{appendix:simplified_bounds} that $\beta_{\max}$ is inversely proportional to $ c$ and $ A$. This implies that decreasing $c$ allows for a wider feasible range of $\beta$. By comparing the accuracies on minority and majority training groups in case $\beta=0, \beta=1$, we can see that both cases have around $100 \%$ accuracy on minority but higher accuracy on majority for $\beta=0$. However, $\beta=1$ results in better generalization overall. This highlights the importance of directing the training process toward a debiased direction early in training, especially when overfitting is likely to occur on the minority group (e.g., when it contains very few samples). 
\paragraph{How the selection of $t$ influences the optimal value of $\beta$.}
To answer this question, we conduct experiments on C-MNIST dataset with $0.5\%$ of minority group. We first train a biased model $\tilde{f}$, and use its predictions to generate the pseudo-attributes for five different values of the threshold $t$. Let's refer to the pseudo-groups with $\tilde{a}=1$ as the pseudo-minority groups. For each threshold, we tune $\beta$ to achieve the best average test accuracy. Simultaneously, we compute the average precision and recall for the minority group.
As shown in \cref{fig:threshold}, with $t =5 \times10^{-4}$, the average precision reaches 1, indicating that all the samples in the pseudo-minority groups are from the minority groups. However, these samples constitute less than 20\% of the total minority groups, as indicated by the average recall. Despite this, GERNE achieves a high accuracy of approximately 70\%, remaining competitive with other methods reported in \cref{table:synth_data} while using only a very limited number of minority samples ($t =5 \times10^{-4}$ corresponds to about 28 samples versus 249 minority samples out of 55,000 samples in the training set). 
As $t$ increases to $10^{-3}$ and $3\times10^{-3}$, precision remains close to 1 while increasing the number of minority samples in the pseudo-minority groups. This increase introduces more diversity among minority samples within the pseudo-minority groups, allowing for lower $\beta$ values to achieve the best average test accuracy. However, for even higher thresholds, such as $t=10^{-2}$, minority samples constitute less than 40\% in the pseudo-minority groups, prompting a need to revert to higher $\beta$ values. We conclude that identifying the samples of minority groups (high average precision and high recall) is of utmost importance for achieving optimal results, and this agrees with the results presented in both \cref{table:synth_data}, \cref{table:real_data}, where we achieve the best results in the known attributes case.

\section{Conclusion}
We introduce GERNE, a novel debiasing approach that effectively mitigates spurious correlations by leveraging an extrapolated gradient update. By defining a debiasing direction from loss gradients computed on batches with varying degrees of spurious correlations, GERNE's tunable extrapolation factor allows optimizing either Group-Balanced Accuracy (GBA) or Worst-Group Accuracy (WGA). Our comprehensive evaluations across vision and NLP benchmarks demonstrate GERNE's superior performance over state-of-the-art methods, both for known and unknown attribute cases, without data augmentation. Furthermore, GERNE offers a general framework that encompasses methods like ERM and Resampling, extending its applicability to unbiased datasets. Future work will explore dynamic adaptation of the extrapolation factor and refine attribute estimation for the unknown attributes case.

\section*{Acknowledgments.} 
This work was funded by the Carl Zeiss Foundation within the project Sensorized Surgery, Germany (P2022-06-004). Maha Shadaydeh is supported by the Deutsche Forschungsgemeinschaft (DFG, German Research Foundation) –  Individual Research Grant SH 1682/1-1.

{
    \small
    \bibliographystyle{ieeenat_fullname}
    \bibliography{main}
}

\clearpage
\maketitlesupplementary
\appendix

\section{The Extrapolated Conditional Attribute Distribution of GERNE:}
\label{appendix:proof_p_ext}
$\mathcal{L}$ in 
\cref{eq:general_loss} can be written as: 
\begin{align}
\mathcal{L} =
\int \ell(y, f(x)) p(y) \, p(a|y) \, p(x|a,y) \, dx \, dy \, da. \notag
\end{align}
Therefore, $\mathcal{L}_{b}, \mathcal{L}_{lb}$ can be written as: 
\begin{equation}
\label{eq:L_b_int}
\mathcal{L}_{b} = \int \ell(y, f(x)) p(y) \, p_{b}(a|y) \, p(x|a,y) \, dx \, dy \, da,
\end{equation}
\begin{equation}
\label{eq:L_lb_int}
\mathcal{L}_{lb} = \int \ell(y, f(x)) p(y) \, p_{lb}(a|y) \, p(x|a,y) \, dx \, dy \, da.
\end{equation}
Where $p_{b}(a|y), p_{lb}(a|y)$ defined in \cref{eq:alphayj}, \cref{eq:rel_ps}, respectively, and $x$ is uniformly sampled within each group $\mathcal{X}_{y, a}$ (i.e., $p(x|y, a) = \frac{1}{|\mathcal{X}_{y, a}|}$).
Substituting \cref{eq:L_b_int}, \cref{eq:L_lb_int} in \cref{eq:relation_losses}:
\begin{align}
    \mathcal{L}_{ext} &= \int \ell(y, f(x)) \, p(y) \, \Big(p_{lb}(a|y) \nonumber \\
    &\quad +\beta \cdot (p_{lb}(a|y)-p_{b}(a|y))\Big) \cdot p(x|a,y) \, dx \, dy \, da \nonumber \\
    & = \int \ell(y, f(x)) \,p(y) \, \Big(\alpha_{ya} + c \cdot (\beta+1) \cdot (\frac{1}{A}-\alpha_{ya}) \Big) \cdot \nonumber \\
    &\quad p(x|a,y) \, dx \, dy \, da \nonumber \\
    & = \int \ell(y, f(x)) \,p(y) \, p_{ext}(a|y) \cdot p(x|a,y) \, dx \, dy \, da.\nonumber
\end{align}
Then: 
\[
p_{ext}(a|y) = \alpha_{ya} + c \cdot (\beta+1) \cdot (\frac{1}{A}-\alpha_{ya}).
\]
Furthermore, we can write $ \mathcal{L}_{ext}$ as follows:
\begin{align*}
\mathcal{L}_{\text{ext}} 
&= \mathbb{E}_{y \sim p(y)} \left[ 
    \mathbb{E}_{a \sim p_{\text{ext}}(a|y)} \left[ 
        \mathbb{E}_{x \sim p(x|a,y)} \left[ \ell(y, f(x)) \right] 
    \right] 
\right] \nonumber \\
&= \frac{1}{K} \cdot \sum_{g=(y,a) \in \mathcal{G}} p_{\text{ext}}(a|y; \beta) \cdot L_g,
\end{align*} with $
L_g = \mathbb{E}_{x \sim p(x| (y, a)=g)}\left[ \ell(y, f(x)) \right]$ by using the discrete expectations over $y$ and $a|y$, with $p(y) = \frac{1}{K}$.

\section{GERNE Versus an Equivalent Sampling and Weighting Approach} 
\label{appendix:GERNE_variance}
We compare GERNE with an equivalent (in terms of loss expectation) sampling+weighting method, which we refer to as ``SW''.
For simplicity, we assume the following:
\begin{enumerate}
    \item A binary classification task where the number of classes equals the number of attributes (i.e., $K = A = 2$).
    \item The attributes are known, and the classes are balanced (i.e., $|\mathcal{X}_{y=1}| = |\mathcal{X}_{y=2}|$).
    \item The majority of samples that hold the spurious correlation in each class are aligned with the class label (i.e., $|\mathcal{X}_{y,a=y}|>|\mathcal{X}_{y,a\neq y}|$), and the dataset is highly biased (i.e.,  $\frac{|\mathcal{X}_{y,a \neq y}|}{|\mathcal{X}_{y,a=y}|} \ll 1$).
    \item In a highly biased dataset, best performance is coupled with overrepresenting the minority groups (as illustrated in \cref{fig:tuning_beta}) in early stages of training. Therefore, an overfitting on the minority groups is expected before the overfitting on the majority groups.
    
\end{enumerate}
We refer to the expected loss of the majority samples as $\mathcal{L}_A$, and the expected loss of the minority as $\mathcal{L}_C$.
For GERNE, we sample two batches: biased and less biased batch, each of size $B$. From \cref{eq:relation_losses}, $\mathcal{L}_{ext}$ can be written as:
\begin{equation}
\label{eq:l_ext_rewrite}
    \mathcal{L}_{ext} = (1+\beta)\cdot \mathcal{L}_{lb}-\beta \cdot \mathcal{L}_b.
\end{equation}
Since the biased batch reflects the inherent bias present in the dataset, under the third assumption, we can approximate $\mathcal{L}_{b}$ by $\mathcal{L}_{A}$, neglecting the loss on the very few samples of the minority group in the batch. Therefore, we have:
\begin{equation}
\label{eq:l_b_approx}
    \mathcal{L}_{b} \approx \mathcal{L}_{A}.
\end{equation}
Following the third assumption and the conditional attribute distribution in \cref{eq:rel_ps}, we can approximate the composition of the less biased batch as follows: a proportion of $(1-\frac{c}{2})$ of the samples in the less biased batch are drawn from the aligned samples (majority group), while a proportion of $\frac{c}{2}$ of the samples from the minority group. This leads to the following approximation:
\begin{equation}
\label{eq:l_lb_approx}
    \mathcal{L}_{lb} \approx (1-\frac{c}{2}) \cdot \mathcal{L}_{A}+\frac{c}{2} \cdot \mathcal{L}_{C}.
\end{equation}
Substituting \cref{eq:l_b_approx}, \cref{eq:l_lb_approx} into \cref{eq:l_ext_rewrite}:
\begin{equation}
    \label{eq:L_ext_b}
    \mathcal{L}_{ext} \approx \frac{2-c \cdot(1+\beta)}{2} \cdot \mathcal{L}_A+\frac{c \cdot(1+\beta)}{2} \cdot \mathcal{L}_C.
\end{equation}
We consider the following ``SW'' approach:
\begin{itemize}
    \item Sampling step: we sample an ``SW'' batch of size $B$, analogous to the less biased batch in GERNE, where $(1-\frac{c}{2})$ of the samples are from the majority group (aligned samples) and $\frac{c}{2}$ from minority group.
    \item Weighting step: we compute the loss $\mathcal{L}_{sw}$ over the sampled batch as follows:
    \begin{equation}
    \label{eq:L_sw_b}
        \mathcal{L}_{sw} = w \cdot \mathcal{L}_A + (1-w) \cdot \mathcal{L}_C, w = \frac{2- c \cdot (1+\beta)}{2},
    \end{equation}
where $\mathcal{L}_A$ is computed over the samples of the majority group in the ``SW'' batch and $\mathcal{L}_C$ is computed over the samples of the minority group.
\end{itemize}
Both $\mathcal{L}_{ext}$ and $\mathcal{L}_{sw}$ in \cref{eq:L_ext_b}, \cref{eq:L_sw_b} are equivalent in expectation.
Let's compute the variance of the two losses:
\begin{align}
 Var(\mathcal{L}_{sw}) &= w^2\cdot Var(\mathcal{L}_A^{1-c/2}) +(1-w)^2 \cdot Var(\mathcal{L}_C^{c/2}) \notag \\
 &\quad + 2\cdot w\cdot(1-w)\cdot Cov(\mathcal{L}_A^{1-c/2}, \mathcal{L}_C^{c/2}),
 \end{align}
where $Var(\mathcal{L}^{m})$ means the variance computed over $m \cdot B$ samples where $B$ is the batch size. For simplicity, we refer to $Var(\mathcal{L}^1) \text{ as } Var(\mathcal{L})$. Following the fourth assumption, when the model overfits on the samples of the minority group (i.e., $\mathcal{L}_C \approx 0$), we can neglect both $Var(\mathcal{L}_C), Cov(\mathcal{L}_A, \mathcal{L}_C)$ terms. Therefore:
\begin{align}
    Var(\mathcal{L}_{sw}) &\approx w^2 \cdot Var(\mathcal{L}_A^{1-c/2})= \frac{w^2}{1-\frac{c}{2}} \cdot Var(\mathcal{L}_A) =\notag \\
    &\quad (\frac{2- c \cdot (1+\beta)}{2})^2\cdot \frac{2}{2-c} \cdot Var(\mathcal{L}_A).
\end{align}
From \cref{eq:l_ext_rewrite}:
\begin{align}
\label{eq:L_ext_inequality}
    Var(\mathcal{L}_{ext}) &= (1+\beta)^2\cdot Var(\mathcal{L}_{lb}) + \beta^2 \cdot Var(\mathcal{L}_b) \notag \\
    &\quad- 2\cdot (1+\beta) \cdot \beta \cdot Cov(\mathcal{L}_{lb}, \mathcal{L}_{b}) \notag \\
    &\quad \geq ((1+\beta) \cdot \sqrt{Var(\mathcal{L}_{lb})}-\beta \cdot \sqrt{Var(\mathcal{L}_b)})^2.
\end{align}
Note that the inequality reduces to an equality in \cref{eq:L_ext_inequality} if $Cov(\mathcal{L}_{lb}, \mathcal{L}_{b}) = \sqrt{Var(\mathcal{L}_{lb})}\cdot \sqrt{Var(\mathcal{L}_b)}$.\\
The covariance term $Cov(.,.)$ can be controlled by the number of shared samples between the biased and less biased batches. If all the aligned samples in the less biased batch are included in the biased batch (i.e., the less biased batch is created by replacing some samples of the majority group with samples from the minority group), we then maximize $Cov(.,.)$. From \cref{eq:l_lb_approx}, we can write:
\begin{align}
\label{eq:var_lb_final}
    Var(\mathcal{L}_{lb}) \approx (1-\frac{c}{2})^2 \cdot Var(\mathcal{L}_A^{1-c/2})=(1-\frac{c}{2}) \cdot Var(\mathcal{L}_A),
\end{align}
and from \cref{eq:l_b_approx}:
\begin{equation}
\label{eq:var_b_final}
    Var(\mathcal{L}_{b}) \approx Var(\mathcal{L}_A).
\end{equation}
Finally, substituting \cref{eq:var_lb_final} and \cref{eq:var_b_final} in \cref{eq:L_ext_inequality}:
\begin{equation}
    Var(\mathcal{L}_{ext}) \geq ((1+\beta)\cdot \sqrt{1-\frac{c}{2}}-\beta)^2 \cdot Var(\mathcal{L}_A).
\end{equation}

According to the fourth assumption, we are interested in the range where $c \cdot (\beta + 1) \geq 1$. Using the limits of $\beta$ defined in \cref{eq:intervals_beta}, we obtain $\beta \in \left[\frac{1-c}{c}, \frac{2-c}{c}\right]$. 
As $\beta \to \frac{2-c}{c}$, the representation of the aligned samples is vanishing (according to \cref{eq:p_ext}) in the sampled batches, which leads to $\mathcal{L}_A > 0$. Assuming a limited and non-vanishing variance $Var(\mathcal{L}_A)$ (i.e., the model outputs non-constant predictions for samples from the majority group), we have:

$\beta \to \frac{2-c}{c} \implies Var(\mathcal{L}_{sw}) \approx 0, \text{ while } Var(\mathcal{L}_{ext}) \neq 0$ for $c \in (0,1]$. This non-vanishing variance of GERNE's loss, if controlled with tuning $\beta$ to ensure stability, gives the model the chance of escaping sharp minima similar to gradient random perturbation \cite{escape_r} and therefore, improves generalization \cite{gradient_noise, keskar2017on, NIPS2017_217e342f}.
\section{\texorpdfstring{Simplifying the Bounds of $\beta$}{Bounding beta}}
\label{appendix:simplified_bounds}
We aim to simplify the upper and lower bounds of $\beta$ in \cref{eq:intervals_beta}. 
We start by simplifying the upper bound: 
\[
\min_{\substack{{(y,a) \in \mathcal{G}} \\ \alpha_{ya} \neq \frac{1}{A}}}\max(i_{ya}^1,i_{ya}^2),
\]
where
$ i_{ya}^1 = -\frac{\alpha_{ya}}{c\cdot (\frac{1}{A}-\alpha_{ya})} -1, i_{ya}^2 = \frac{1-\alpha_{ya}}{c\cdot(\frac{1}{A}-\alpha_{ya})} -1 $, and under the following constraints: $\forall y \in \mathcal{Y}, \sum_a \alpha_{ya} = 1$, $\alpha_{ya} \in ]0,1[\setminus \{\frac{1}{A}\}$, $A \ge 2$, and $c \in (0,1]$.

We note that $i_{ya}^1$ is a decreasing, and $i_{ya}^2$increasing function in $\alpha_{ya}$. We can show that if $\alpha_{ya} < \frac{1}{A}$, then $i_{ya}^2 > i_{ya}^1$, and if $\alpha_{ya} > \frac{1}{A}$, then $i_{ya}^1 > i_{ya}^2$.
We conclude with the following: 
\[
\min_{\substack{{(y,a) \in \mathcal{G}} \\ \alpha_{ya} < \frac{1}{A}}}\max(i_{ya}^1,i_{ya}^2) = i_{y'a'}^2, \min_{\substack{{(y,a) \in \mathcal{G}} \\ \alpha_{ya} > \frac{1}{A}}}\max(i_{ya}^1,i_{ya}^2) = i_{y''a''}^1, 
\]
where: 
\[\alpha_{y'a'} = \min_{\substack{{(y,a) \in \mathcal{G}} }}\alpha_{ya}<\frac{1}{A}, \alpha_{y''a''} = \max_{\substack{{(y,a) \in \mathcal{G}} }}\alpha_{ya}>\frac{1}{A}.\]

Since $\sum_k \alpha_{y'k} = 1$, we have:
\[
\sum_{k \ne a'} \alpha_{y'k} = 1 - \alpha_{y'a'},
\]
which implies that there exists $j \ne a'$ such that:
\[
\alpha_{y'j} \ge \frac{1 - \alpha_{y'a'}}{A - 1}.
\]
Given that $\alpha_{y'a'} < \frac{1}{A}$ and $A \ge 2$, it follows that 
\[\frac{1 - \alpha_{y'a'}}{A - 1} > \frac{1}{A},\]
and hence $\alpha_{y'j} > \frac{1}{A}$. Therefore, we have 
\[\max(i_{y'j}^1,i_{y'j}^2) = i_{y'j}^1 = -\frac{\alpha_{y'j}}{c\cdot (\frac{1}{A}-\alpha_{y'j})} -1.
\]
Since $i_{ya}^1$ is a decreasing function in $\alpha_{ya} \in (\frac{1}{A}, 1]$, we have 
\[i_{y'j}^1 \leq -\frac{\frac{1-\alpha_{y'a'}}{A-1}}{c \cdot (\frac{1}{A}-\frac{1-\alpha_{y'a'}}{A-1})}-1 = \frac{1-\alpha_{y'a'}}{c \cdot (\frac{1}{A}-\alpha_{y'a'})}-1 = i_{y'a'}^2.\]
Thus: $i_{y'j}^1 \leq i_{y'a'}^2$, and since $i_{y''a''}^1 \leq i_{y'j}^1$, we conclude that the upper bound of $\beta$ is: $\beta_{\max} = i_{y''a''}^1$. 
For the lower bound of $\beta$, we can follow the same previous step, but we choose $\beta = -1$ as the lower bound ($\beta = -1$ satisfies \cref{eq:intervals_beta} and simulates ERM training as shown in \cref{sec:general_framework}). In conclusion, for the known attributes case, we tune $\beta$ within the interval: $ [\beta_{\min}, \beta_{\max}]= [-1, i_{y''a''}^1]$. 
The upper bound $\beta_{\max}$ is inversely proportional to both $c, A$. Consequently, larger values of $c$ reduce the feasible range for the extrapolation factor, making GERNE appear more sensitive to small variations in $\beta$.
\section{Algorithm 2}
\label{appendix:algorithm2}
\label{appendix:algo2.}
\begin{algorithm}[H]
\caption{GERNE for the unknown attribute case}
\label{alg:unknown_attr_algo}
\textbf{Input:} $\mathcal{X}_{y} \subseteq \mathcal{X}$ for $ y \in \mathcal{Y} $, $f$ with initial $\theta = \theta_0 , \tilde{\theta} = \tilde{\theta_0}$ (parameters of the biased model $\tilde{f}$), $\#$ epochs $E$, batch size per class label $B$, $\#$ classes $K$, $\#$ attributes $\tilde{A} = 2$, learning rate $\eta$.
\begin{algorithmic}[1]
\State Training $\tilde{f}$ on biased batches with class balanced accuracy CBA $= \frac{1}{K} \sum_{y \in \mathcal{Y}} \mathbb{P}_{x|y}(y = \arg \max_{y' \in \mathcal{Y}} \tilde{f}_{y'}(x)) \quad$ as the evaluation metric for model selection.
\State Select a threshold $t$ and create the pseudo-groups $\tilde{\mathcal{G}}$: For each class $y$, we compute the predictions $\tilde{y}_i = \text{softmax}(\tilde{f}(x_i))_y$ for each $x_i \in \mathcal{X}_y$. We then form the pseudo-minority group $\mathcal{X}_{y, \tilde{a}=1}$ as the samples $x_i$ that have the smallest \( \lfloor t \cdot |\mathcal{X}_y| \rfloor \) predictions. The remaining samples form the pseudo-majority group $\mathcal{X}_{y, \tilde{a}=2} = \mathcal{X}_y \setminus \mathcal{X}_{y, \tilde{a}=1} $.

\State Follow \textbf{Algorithm 1} with $\tilde{\mathcal{G}}$ replacing $\mathcal{G}$. We consider a higher upper bound for $\beta$ than $\beta_{\max}$ derived in \cref{appendix:simplified_bounds} as justified by the proof of Proposition 1. in \cref{appendix:useful_beta}.
\end{algorithmic}
\end{algorithm}

\section{Proposition 1.}
\label{appendix:useful_beta}
Creating both biased and less biased batches using the pseudo-groups $\mathcal{\tilde{G}}$, and with $\beta$ as a hyperparameter, we can simulate batches with a more controllable conditional attribute distribution. Specifically, for $(y,a) \in \mathcal{G}$, we can achieve scenarios where $p_{ext}(a|y) > \max_{\tilde{a} \in \mathcal{\tilde{A}}}p(a|\tilde{a},y) \text{ or } p_{ext}(a|y) < \min_{\tilde{a} \in \mathcal{\tilde{A}}}p(a|\tilde{a},y)$ as opposed to \cref{eq:impossible_cond}.\\
\textbf{Proof.}
We define $\alpha_{y\tilde{a}}$ the same way as in \cref{eq:alphayj} for the created pseudo-groups: $\alpha_{y\tilde{a}} = \frac{|\mathcal{X}_{y, \tilde{a}}|}{|\mathcal{X}_{y}|}$.
For a constant $c$ and $A=2$, we create the less biased batch as in \cref{eq:rel_ps}:
\begin{equation}
\label{eq:p_lb_app}
    p_{lb}(\tilde{a}|y)=\alpha_{y\tilde{a}}+c\cdot (\frac{1}{2}-\alpha_{y\tilde{a}}).
\end{equation}
Similar to \cref{eq:p_ext}, the conditional attribute distribution $p_{ext}(\tilde{a}|y)$ is given by:
\begin{align}
\label{eq:new_pd}
p_{ext}(\tilde{a}|y)= \alpha_{y\tilde{a}}+ c\cdot (\beta+1) \cdot(\frac{1}{2}-\alpha_{y\tilde{a}}).
\end{align}
We can write $p_{ext}(a|y) $ as follows:
\begin{equation}
\label{eq:new_pd_a}
    p_{ext}(a|y) = \sum_{\tilde{a} \in \tilde{\mathcal{A}}} p_{ext}(\tilde{a}|y)\cdot p(a|\tilde{a},y).
\end{equation}
Placing \cref{eq:new_pd} in \cref{eq:new_pd_a}, we get
\begin{equation}
\label{eq:final_p_d}
    p_{ext}(a|y) = \sum_{\tilde{a} \in \tilde{\mathcal{A}}} \alpha_{y\tilde{a}} \cdot p(a|\tilde{a},y) + c\cdot(\beta+1)\cdot(\frac{1}{2}-\alpha_{y\tilde{a}})\cdot p(a|\tilde{a},y).
\end{equation}
For $p(a|\tilde{a}=1,y) \neq p(a|\tilde{a}=2,y)$ and $\alpha_{y1} \neq \frac{1}{2}$, to make $p_{ext}(a|y) = p $ for some $p \in [0,1]$, we tune $\beta$ until reaching the $\beta_{target}$ defined as follows:
\begin{equation}
    \beta_{target} = \frac{p-\sum_{\tilde{a} \in \tilde{\mathcal{A}}} \alpha_{y\tilde{a}}\cdot p(a|\tilde{a},y)}{\sum_{\tilde{a} \in \tilde{\mathcal{A}}}c\cdot(\frac{1}{2}-\alpha_{y\tilde{a}})\cdot p(a|\tilde{a},y)}-1.
\end{equation}

\textbf{Discussion.} When $\alpha_{y1} = \frac{1}{2}$, our algorithm is equivalent to sampling uniformly from $\mathcal{X}_y$ and equally from classes. When $p(a|\tilde{a}=1,y) = p(a|\tilde{a}=2,y)$, it implies that  $\tilde{f}$ has distributed the samples with attribute $a$ and class $y$ equally between the two pseudo-groups $\mathcal{X}_{y,\tilde{a}\in \mathcal{A}}$. However, in practice, this is exactly the scenario that $\tilde{f}$ is designed to avoid. Specifically, if $a$ represents the presence of spurious attributes (i.e., the majority group), it is likely that $p(a|\tilde{a}=1,y) < p(a|\tilde{a}=2,y)$. Conversely, when $a$ represents the absence of spurious features (i.e., the minority group), we would expect $p(a|\tilde{a}=1,y) > p(a|\tilde{a}=2,y)$. In fact, $\tilde{f}$ is explicitly trained to exhibit a degree of bias, which inherently disrupts the above equality.

\section{Implementation Details}
\subsection{Implementation Details for Datasets-1}
\label{appendix:baselines_impdetails}
For the C-MNIST \cite{arjovsky2020invariantriskminimization, lee_learning_2021}, we deploy a multi-layer perceptron (MLP) with three fully connected layers, while for C-CIFAR-10 \cite{hendrycks2019benchmarking, LfF} and bFFHQ \cite{karras2019style, lee_learning_2021}, we employ ResNet-18 model \cite{resnet}, pretrained on ImageNet1K \cite{imagenet}, as the backbone. We apply the Stochastic Gradient Descent (SGD) optimizer with a momentum of 0.9 and a weight decay of $10^{-2}$ across all three datasets. 
We set the batch size to 100 per group or pseudo-group for both C-MNIST and C-CIFAR-10, and to 32 for bFFHQ. For C-MNIST, we use a learning rate of $10^{-1}$ in the known attributes case and $10^{-2}$ in the unknown attributes case. For C-CIFAR-10 and bFFHQ, we use a learning rate of $10^{-4}$.
 
In the unknown attribute case, we treat the threshold $t$ as an additional hyperparameter that requires tuning. 
We avoid using any data augmentations, as certain transformations can unintentionally fail to preserve the original label. For example, flips and rotations in C-MNIST can distort labels (e.g., a rotated ``6'' appearing as a ``9'') \cite{shorten2019survey}. 
For training $\tilde{f}$ in case of unknown attributes in the training set, we employ the same model architecture as $f$, with modifications to the hyperparameters: the weight decay is doubled, and the learning rate is reduced to one-tenth of the learning rate used to train $f$. The loss function used is the Cross-entropy loss for all the experiments.
\subsection{Implementation Details for Datasets-2}
\label{appendix:baselines_impdetails_dataset2}
To ensure a fair comparison between GERNE and the methods in \cite{ChangeIsHard}, we adopt the same experimental settings. For Waterbirds \cite{waterbirds} and CelebA \cite{celeba} datasets, we use a pretrained ResNet-50 model \cite{resnet} as the backbone, while for CivilComments \cite{civilcomments}, we use a pretrained BERT model \cite{bert}. We append an MLP classification head with $K$ outputs. 
We employ SGD with a momentum of 0.9 and a weight decay of $10^{-2}$ for Waterbirds and CelebA. For CivilComments, we use AdamW \cite{Kingma2015AdamAM} optimizer with a weight decay of $10^{-4}$ and a tunable dropout rate. We set batch sizes to 32 for both Waterbirds and CelebA and 5(16) per group(pseudo-group) for CivilComments. The learning rates are configured as follows: $10^{-4}$ for Waterbirds and CelebA, and $10^{-5}$ for CivilComments. Additionally, we set the bias reduction factors $c$ to 0.5 for Waterbirds and CelebA and to 1 for CivilComments.
For image datasets, we resize and center-crop the images to 224×224 pixels. In the case of unknown attributes in the training set, $\tilde{f}$ has the same architecture as $f$, but we adjust the hyperparameters: the weight decay is doubled, and the learning rate is reduced to one-tenth of the value used to train $f$. We employ the Cross-entropy loss as the loss function across all experiments. For experiments with unknown attributes in both the training and validation sets, we limit the search space for $t$ to the interval $[0,\frac{1}{2}]$.

\section{Evaluating GERNE with Limited Attribute Information}
\label{appendix:GERNE_limited_info}
To further demonstrate the effectiveness of GERNE in scenarios with limited access to samples with known attributes, we conduct two experiments on the CelebA dataset using only the validation set with its attribute information for training (excluding the training set). We follow the same experimental setup and implementation details outlined in \cref{appendix:baselines_impdetails_dataset2}. As part of the implementation, we first tune the hyperparameters using the designated evaluation metric. Once we determine the optimal hyperparameters, we fix them and train the model $f$ three times with different random seeds. Finally, we report the average worst-group test accuracy and standard deviation across these runs.

\textbf{Evaluation on the Test Set.} In this experiment, we train the model $f$ using the entire validation set and use the worst-group test accuracy as the evaluation metric. This setup represents the best possible performance achievable when relying solely on the validation set for training. 

\textbf{Cross-Validation.} 
In this experiment, we divide the validation set into three non-overlapping folds, ensuring that each fold preserves the same group distribution as the original validation set. Specifically, we randomly and evenly distribute samples from each group in the validation set across the folds. We train $f$ using two of the folds, and use the remaining fold for hyperparameter tuning and model selection, where the worst-group accuracy on this fold serves as the evaluation metric. We repeat this process three times so that each fold is used once as the validation fold. Finally, we report the average worst-group test accuracy and standard deviation across all nine runs (three folds × three seeds) in \cref{tab:valid_train_celeba}.

We compare the results of GERNE with DFR, a method that trains the final layer on the validation set following ERM training on the training set. GERNE consistently achieves state-of-the-art results, demonstrating its robustness and effectiveness even under limited attribute information.
\begin{table}[h!]
\centering
\caption{Performance comparison of GERNE and DFR. GERNE uses only the validation set for training. We report the worst-group test accuracy (\%) and standard deviation over three trials.}
\label{tab:valid_train_celeba}
\small
\begin{tabular}{lc}
\toprule
\textbf{Method} & \textbf{WGA on Test Set (\%)} \\
\midrule
DFR & 86.30 $\pm$ 0.30  \\
GERNE --- Evaluation on Test Set & 90.97 $\pm$ 0.35 \\
GERNE --- Cross-Validation & 88.63 $\pm$ 0.59 \\
\bottomrule
\end{tabular}
\end{table}

\end{document}